%% file: ArXiv.tex
\documentclass[10pt,twocolumn,letterpaper]{article}

\usepackage{iccv}
\usepackage{times}
\usepackage{epsfig}
\usepackage{graphicx}
\usepackage{amsmath}
\usepackage{amssymb}
\usepackage{booktabs}
\usepackage{multirow}
\usepackage{animate}
\usepackage{inconsolata}
\usepackage[pagebackref=true,breaklinks=true,letterpaper=true,colorlinks,bookmarks=false]{hyperref}

\usepackage[pagebackref=true,breaklinks=true,letterpaper=true,colorlinks,bookmarks=false]{hyperref}

\iccvfinalcopy 

\usepackage[colorinlistoftodos,bordercolor=orange,backgroundcolor=orange!20,linecolor=orange,textsize=scriptsize]{todonotes}
\usepackage[utf8]{inputenc}

\newcommand{\jaccard}{$\mathcal{J}$ }
\newcommand{\fmax}{$\mathcal{F}$ }
\newcommand{\LTA}{Long-term Attention }
\newcommand{\STA}{Short-term Attention }
\newcommand{\trainVal}{\textit{Train33 }}

\def\iccvPaperID{1034} 

\ificcvfinal\fi
        
\begin{document}

\title{MAIN: Multi-Attention Instance Network for Video Segmentation}

\author{ \normalsize Juan León Alcázar$^{\ast 1}$, María A. Bravo$^{\ast 1}$, Ali K. Thabet$^{2}$, Guillaume Jeanneret$^{1}$, Thomas Brox$^{3}$,  \\ \normalsize Pablo Arbeláez$^{1}$  and Bernard Ghanem$^{2}$ \\ 
\normalsize $^{1}$Universidad de los Andes, $^{2}$King Abdullah University of Science and Technology, $^{3}$University of Freiburg \\
\small $^{1}$\{\texttt{jc.leon, ma.bravo641, g.jeanneret10, pa.arbelaez}\}\texttt{@uniandes.edu.co;} \\
\small $^{2}$\{\texttt{ali.thabet, bernard.ghanem}\}\texttt{@kaust.edu.sa}; \\ \small $^{3}$\texttt{brox@cs.uni-freiburg.de}
}


\maketitle
\let\thefootnote\relax\footnotetext{$^{\ast}$ Equal contribution}

\begin{abstract}
\vspace{-0.30cm}
   Instance-level video segmentation requires a solid integration of spatial and temporal information. However, current methods rely mostly on domain-specific information (online learning) to produce accurate instance-level segmentations. We propose a novel approach that relies exclusively on the integration of generic spatio-temporal attention cues. Our strategy, named \emph{Multi-Attention Instance Network (MAIN)}, overcomes challenging segmentation scenarios over arbitrary videos without modelling sequence- or instance-specific knowledge. We design MAIN to segment multiple instances in a single forward pass, and optimize it with a novel loss function that favors class agnostic predictions and assigns instance-specific penalties. We achieve state-of-the-art performance on the challenging Youtube-VOS dataset and benchmark, improving the unseen Jaccard and F-Metric by 6.8\% and 12.7\% respectively, while operating at real-time (30.3 FPS). 
   \vspace{-0.50cm}
   
\end{abstract}

\input{sections/1Introduction}

\input{sections/2RelatedWork}

\input{sections/3Method}

\input{sections/4Experiments}
\input{sections/5ComparisonSA}

\input{sections/6QualitativeConclusions}
\input{sections/7Acknowledgement}

{\small
\bibliographystyle{ieee}
\bibliography{egbib}
}
\pagebreak
\input{sections/8Apendix}
\end{document}

%% file: sections/1Introduction.tex
\vspace{-0.10cm}
\section{Introduction}
\vspace{-0.10cm}
Current state-of-the-art video segmentation methods \cite{caelles2016one, voigtlaender2017online, khoreva2017lucid, Li_2018_ECCV} have achieved impressive results for the binary task of separating foreground objects from the background. However, the finer-grained task of multi-instance video segmentation, which aims at independently identifying and segmenting multiple objects, remains an open research problem. There are several task-specific challenges for multi-instance segmentation. First, an accurate label assignment must create a set of spatial and temporal consistent masks across all the instances in the sequence. Second, there is a loose definition (lack of semantics) for the object of interest, since segmentation targets are arbitrarily chosen among all the available objects in the starting frame. Third, the appearance, scale, and visibility of segmentation targets vary throughout the video. Finally, there are complex scene dynamics specific to each sequence, which are hard to model without domain-specific knowledge. Figure \ref{fig:task} shows video clips with some results of our method on such challenging scenarios.

\begin{figure}
    \centering
    \animategraphics[loop, autoplay,width=0.48\textwidth]{8}{animatedImages/images-}{0}{53}
    \caption{\textbf{MAIN video segmentation results.} These videos from the Youtube-VOS dataset present several challenges: large visual similarity, overlap and direct interaction between instances, and high variability in viewpoint and visual appearance. To visualize the animated figure, use Adobe Acrobat Reader. 
    \vspace{-0.50cm}}
    \label{fig:task}
\end{figure}


\begin{figure*}[t!]
    \begin{center}
        \includegraphics[width=0.90\textwidth]{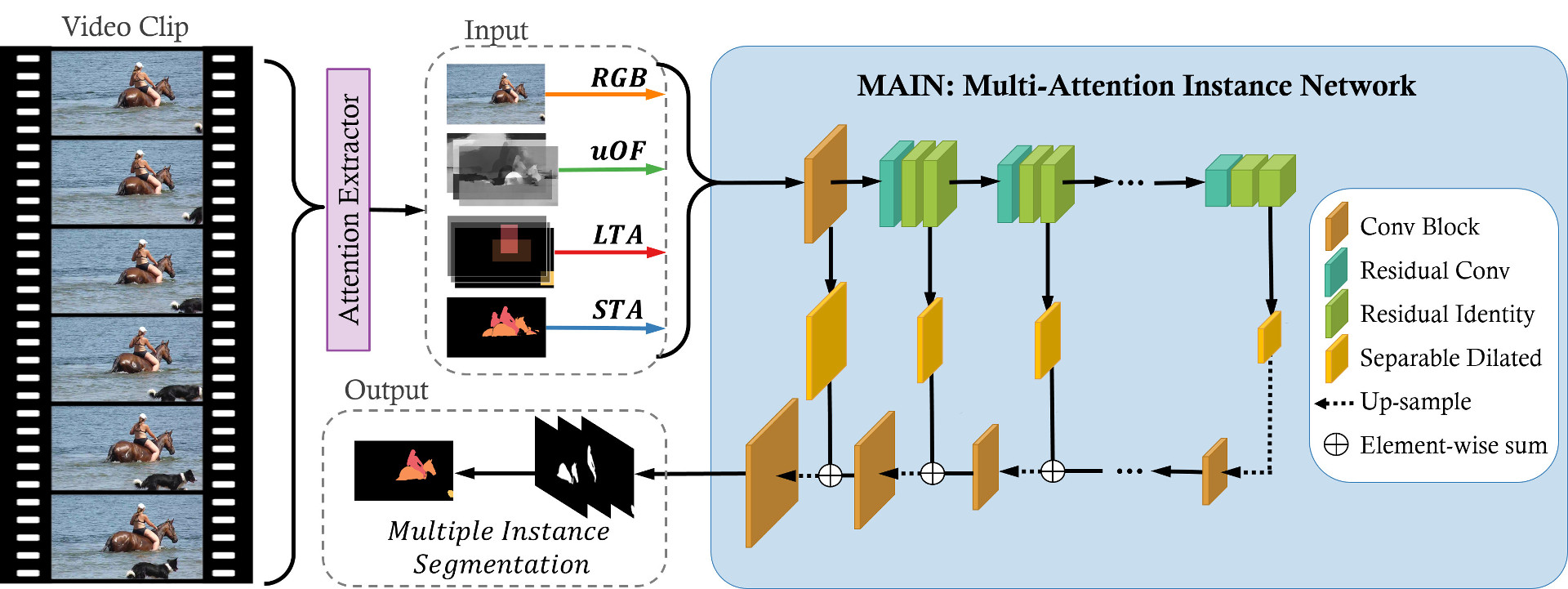}
    \end{center}
    \caption{\textbf{Multi-Attention Instance Network (MAIN).} Our architecture integrates static cues, motion cues, long and short temporal cues in a single network to produce a multi-instance segmentation. 
    We take as input the current RGB image, the Optical Flow (a robust version of the optical flow), and the Long and Short spatio-temporal attention (LTA and STA) cues and use them as input to our network. The decoder in MAIN uses multiple side-outputs that are combined following a Feature Pyramid Architecture of separable dilated convolutions. 
    \vspace{-0.70cm}
    }
    \label{fig:NetworkOverview}
\end{figure*}

One-shot video segmentation defines the task of segmenting arbitrary objects given a ground-truth annotation in the first video frame. Methods like \cite{caelles2016one, voigtlaender2017online, khoreva2017lucid} approached this task by training a binary model over fully annotated videos (a phase commonly known as \textit{Offline Training}) and finetuning it on every unseen video, resulting in multiple sequence-specific or instance-specific models (a phase also known as \textit{Online Training}). These methods rely strongly on online training to estimate and fuse instance-level predictions. Recent methods have proposed other strategies to improve the online phase by using instance re-identification \cite{shaban2017multiple, li2018video, cheng2018fast}, instance proposals \cite{yang2018efficient, hu2017maskrnn, shin2017pixel}, or local motion information \cite{tokmakov2017learning}, among many others. Regardless of the selected strategy, these methods remain computationally expensive for training and inference, and might not be suitable for modern large-scale datasets. 
%

In this paper, we propose a single encoder-decoder architecture that  operates by design at the instance-level in an offline configuration, enabling efficient and generic processing of unconstrained video sequences. Our method focuses on recurrent estimation and consistent propagation of \textit{generic spatio-temporal information}. It integrates static grouping features (\emph{e.g.} color, shape and texture), with short-term motion cues (\emph{e.g.} optical flow), along with short and long spatio-temporal attention mechanisms, which attend to a specific set of objects in a frame and maintain large contextual information. We optimize our model end-to-end using a novel loss function that assigns a pixel-wise penalty according to the relative error contribution of the segmentation target. Figure \ref{fig:NetworkOverview} shows an overview of our method, which we call \textit{Multi-Attention Instance Network} (MAIN).


Intuitively, MAIN incorporates insights from multi-target tracking, image warping, and static image segmentation in an efficient framework that runs at $30.3$ fps. 
To the best of our knowledge, MAIN is the first method that can generate an arbitrary number of instance-level predictions in a single forward pass and without domain-specific information, thus, making it particularly useful for unconstrained large-scale scenarios. MAIN accomplishes state-of-the-art performance and improves with a high increase for unseen instances in the Jaccard and F-Metric to 55.0\% and 63.0\% respectively. We verify our contributions through a series rigorous ablation studies in which we demonstrate the contribution of each additional feature and the robustness and effectiveness of our method.

\vspace{-0.40cm}
\paragraph{Contributions.}
Our work has four main contributions. \textbf{(1)} MAIN directly addresses the multi-instance scenario; it generates multi-instance segmentations in a single forward pass, without the need of domain-specific knowledge or instance-specific fine-tuning. \textbf{(2)} We introduce a novel loss function devised for the multi-instance segmentation scenario. This function enables us to work on large and highly imbalanced datasets that contain multiple instances of diverse sizes without any hyper-parameter tuning. \textbf{(3)} MAIN explicitly fuses static, motion, short-term and long-term temporal grouping cues into a single end-to-end trainable architecture. \textbf{(4)} We propose a dilated separable decoder architecture, which allows us to aggregate multi-scale information using less computationally expensive operations.

To ensure reproducible results and to promote future research, all the resources of this project --source code, model weights, and official benchmark results-- will be made publicly available.

%% file: sections/2RelatedWork.tex
\vspace{-0.10cm}
\section{Related Work}
\label{sec:related_work}
\vspace{-0.10cm}
Interest in video segmentation has grown in recent years within the computer vision community, in part due to the availability of new datasets \cite{perazzi2016benchmark, pont20172017, xu2018youtubeArxiv}, and the development of deep convolutional neural architectures \cite{hinton1989connectionist, lecun1989backpropagation, krizhevsky2012imagenet, hochreiter1997lstm, long2015fully}. But mostly due to the recent re-formulation of the problem, evolving  
from a direct extension of classic image segmentation  \cite{brostow2009semantic, sundberg2011occlusion, galasso2013unified} into a one-shot learning task \cite{perazzi2016benchmark, pont20172017, xu2018youtubeArxiv}, which further highlights the need for temporally consistent estimations. 

\vspace{-0.40cm}
\paragraph{One-Shot learning for video object segmentation.}
Recent datasets formulate the video object segmentation task as a one-shot problem \cite{perazzi2016benchmark, pont20172017, xu2018youtubeArxiv} and provide the ground-truth instance segmentations for the first frame. State-of-the-art methods \cite{caelles2016one, perazzi2017learning, yang2018efficient, voigtlaender2017online, xiao2018monet} usually train a binary offline model, capable of producing an initial background and foreground estimation. Then, during online training, they fine-tune these models for the specific instances in validation using the available ground-truth. Most offline methods \cite{caelles2016one, tokmakov2016weakly, voigtlaender2017online, khoreva2017lucid} do not work in a multi-instance scenario and  require either multiple online training sessions or instance-specific knowledge \cite{yang2018efficient} to do so. In contrast, MAIN works directly on a multi-instance offline configuration, generating multi-instance segmentations in a single forward pass without any prior knowledge or online training. 



\vspace{-0.40cm}
\paragraph{Loss functions for video segmentation.}
Video object segmentation datasets have an inherent large class imbalance favoring trivial background assignments and making small objects into hard false negatives. To tackle this issue, strategies like the Focal Loss \cite{lin2017focal} use a sample-specific weight enabling the large imbalance to be controlled on detection tasks. Likewise, other methods  \cite{xie2015holistically, deng2018learning, maninis2016convolutional} create an adaptive loss function for edge detection weighted by the ratio of positive edge pixels to background pixels. In the segmentation field, Milletari \textit{et al.} \cite{milletari2016v} proposed the Dice Loss function, which directly approximates the Jaccard Index. In this paper, we propose a novel loss function, the Weighted Instance Dice (WID) loss, that exploits the benefits of the Dice loss to perform effectively on highly imbalanced datasets. WID independently weighs every pixel prediction according to the size of the corresponding instance.



\vspace{-0.40cm}
\paragraph{Long-term temporal cues.}
Building upon the core ideas of generic and specific networks, some recent works \cite{tokmakov2016weakly, tokmakov2017learning, xiao2018monet}  tackle the lack of consistent temporal information, a phenomenon that arises when video frames are processed independently. These approaches focus mostly on extracting useful temporal information from moving objects. In fact, motion is a strong bottom-up cue for video segmentation, especially when objects of interest have independent motion patterns \cite{Keuper_2015_ICCV}. Following this line of thought, some approaches \cite{brox2010object, wang2017super} use long-term point trajectories based on dense optical flow fields. They define pair-wise distances between these trajectories and cluster them to have temporally consistent segmentations of moving objects. We build upon these core ideas and design an encoder-decoder architecture that incorporates long-term spatial attention cues estimated from a tracking algorithm.

\vspace{-0.40cm}
\paragraph{Short-term temporal cues.}
Recent video segmentation methods directly rely on motion information, thus including it as prior knowledge \cite{khoreva2017lucid, hu2017maskrnn, perazzi2017learning, hu2018motion, xiao2018monet, vorobyovunsupervised}. These strategies either rely on dense optical flow estimation or another online approximation of pixel-level short-term motion patterns, either pre-computed or estimated jointly with the segmentation \cite{cheng2017segflow}. Compared to these methods, MAIN benefits from motion information by explicitly fusing a robust version of the optical flow with the standard RGB input.

Another important source of information when performing a recurrent task like video segmentation is the set of previous predictions. Methods like VPN \cite{Jampani_2017,cheng2018fast} and MaskTrack \cite{perazzi2017learning} use previously segmented masks to better estimate the current segmentation. We also use the optical flow to warp the previous segmentation and to set our short spatio-temporal prior information. MAIN fuses these short-term cues with long spatio-termporal cues to produce a set of temporally consistent instance segmentations. 

%% file: sections/3Method.tex
\vspace{-0.10cm}
\section{Multi-Attention Instance Network (MAIN)}
\vspace{-0.10cm}


Video object segmentation datasets such as Youtube-VOS \cite{xu2018youtubeArxiv} and DAVIS-17 \cite{pont20172017} contain multiple segmentation targets in a single video. Our approach directly addresses the multi-instance scenario with an offline trained network that produces multiple segmentations in a single forward pass. We concatenate attention sources for every instance in the video and optimize MAIN in order to output a segmentation only if the attention cues indicate the presence of an instance. Given a dataset with at most $N$ instances per video, we use $2M$ attention priors for $M$ possible instances (one short-term (STA) and one long-term (LTA) for each instance), and set the output of our decoder to return a tensor of dimensions $N\times H \times W$. Then, if a video has $M\leq N$ instances, MAIN predicts instances only for the first $M$ channels.


While simple, this extension overcomes two important challenges of the one-shot multi-instance segmentation scenario: (i) the lack of semantics for target objects, since attention cues are class agnostic and (ii) the need for domain-specific knowledge to achieve temporally consistent instance-level segmentations, given by the multiple channel output that generates multi-instance segmentation in a single forward pass. By using this strategy, MAIN reduces the computational complexity of the forward pass, along with the total training time.



\vspace{-0.40cm}
\paragraph{Instance Shuffle.} 
We observe that the instance distribution in Youtube-VOS (Avg. 1.71, Std. 0.87, Mode 1, Max 6 per video) makes it difficult for the network to generate accurate multi-instance segmentations for videos with more than three instances, since they are not frequent. We address this problem by randomly shuffling the attention channels, and performing the same permutation in the output maps and supervision data. After this modification, instances appear with similar frequency across output channels in a single batch regardless of the bias in the dataset.

\vspace{-0.10cm}
\subsection{Weighted Instance Dice (WID) Coefficient Loss}
\label{subsec:instanceDiceCoeff}
\vspace{-0.10cm}
Most state-of-the-art methods for video object segmentation use either a binary or a multi-class cross-entropy loss function that can, optionally, be weighted in the presence of a large class imbalance \cite{maninis2016convolutional,xie2015holistically}. Since MAIN predicts instance-level and class-agnostic segmentations, we depart from the standard practice and introduce a novel loss function that better approximates the multi-instance segmentation scenario. This loss penalizes overlapping predictions and errors over small instances, which have large influence on the evaluation metric. We propose the Weighted Instance Dice Coefficient (\textit{WID}) loss function for a multi-instance segmentation scenario:

\vspace{-0.30cm}
\begin{equation}
    	\textit{WID}(\mathcal{P},\mathcal{G})= \sum^{n}_{i} \alpha (g_{i})(1-D(p_{i},g_{i}))+ \sum^{n}_{i}\sum^{n}_{j \neq i}D(p_{i},p_{j})
\end{equation}
\vspace{-0.30cm}

\noindent where $\mathcal{P} = \{p_{0},p_{1}...,p_{n}\}$ is the set of instance predictions and $\mathcal{G} = \{g_{0},g_{1}...,g_{n}\}$ the set of instance ground-truth. $D$ is the standard Dice coefficient, hence $(1-D(p_{i},g_{i}))$ corresponds to a non-weighted Dice loss for a single instance. $\alpha(g_{i})$ is an instance specific weight coefficient that increases for smaller instances. We define $\alpha(g_{i}) = 1- \frac{|g_i |}{WH}$, where $|g_i|$ is the total number of pixels in $g_i$ and $(W,H)$ correspond to the width and height of the video frame. Finally, $\sum^{n}_{i}\sum^{n}_{j \neq i}D(p_{i},p_{j})$ enforces a penalty for overlapping instance predictions reducing incorrect instance assignments.

Formally, WID is supported by the bijection between the Dice coefficient and the Jaccard index ($J=\frac{D}{2-D}$), which guarantees that minimizing a Dice loss function maximizes the associated Jaccard metric. In contrast to the Dice coefficient proposed by \cite{milletari2016v}, whose convergence rate is modulated by $\frac{2}{(2-D)^2}$, WID presents an improvement as it considers instance overlapping errors and allows for an instance-specific weighting instead of the standard practice of class-specific weighting.

\subsection{Attention Priors}
\begin{figure}[t!]
    \begin{center}
        \includegraphics[width=0.44\textwidth]{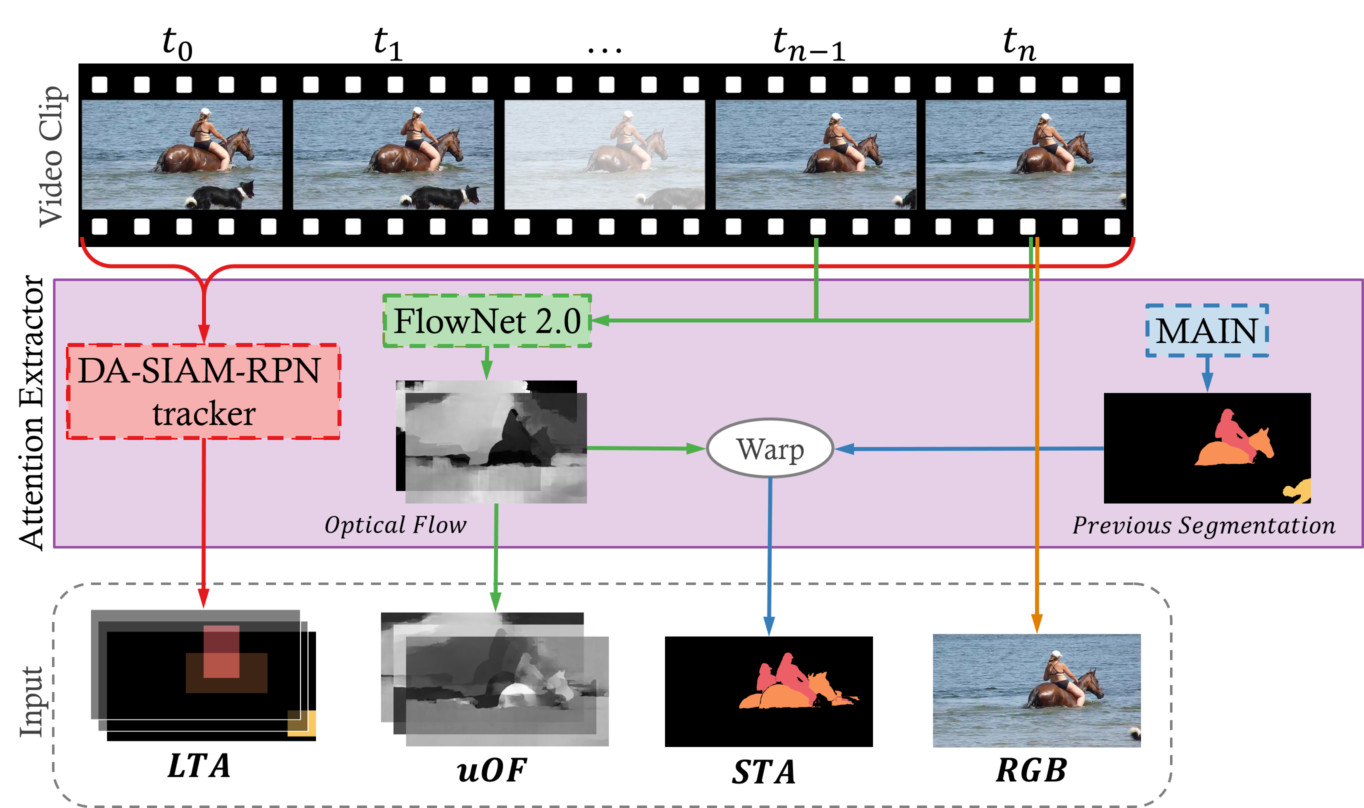}
    \end{center}
    \caption{\textbf{Attention cues.} MAIN integrates multiple attention cues to create temporally consistent instance-level segmentations. During a forward pass of frame $t_n$, we calculate the \LTA (LTA) cue, as a stack of bounding boxes of the target objects, calculated using DA-SIAM-RPN tracker; the Unit Optical Flow (uOF) using FlowNet2.0; and the \STA (STA) cue with the segmentation at time $t_{n-1}$ produced by MAIN warped by the corresponding optical flow.
    \vspace{-0.50cm} 
    }
    \label{fig:Cues}
\end{figure}

\vspace{-0.10cm}
At the core of our approach is a set of attention cues ${A_{m,t}}$ estimated for all $m$ instances at time $t$ given their visual appearance and location at time $t-n$. Initially, we estimate long-term dependencies (\textit{i.e.} $n>>1$), by means of a tracking strategy that creates $m$ attention regions according to the temporal evolution of $n$ previous positive detections. We call these cues \LTA (LTA). We complement these long term recurrences with attention cues over short-term intervals (\textit{i.e.} $n=1$). We estimate a robust version of Optical Flow and use it as a fine-grained (pixel-level) short-term motion attention cue. We also propagate the prediction at time $t-1$ by warping it with the estimated optical flow, creating a coarser (region-level) \STA (STA). Figure \ref{fig:Cues} presents an overview of the selected cues.


\vspace{-0.40cm}
\paragraph{Short-Term Attention Prior.} 
We propose an STA source by including information from the warped segmentation mask of the previous frame. This prior is motivated by the video's causal nature, in which the predicted mask at time $t$ is strongly related to the prediction at time $t-1$. Such temporal correspondence approximates both the target specification and its immediate trajectory. 

We follow the standard definition of a recurrent warping $w(S_t,\mathbf{o}_t)=S_{t+1}$ as the mapping $w:\mathbb{R}^{W\times H\times C} \times \mathbb{R}^{W\times H\times 2} \rightarrow \mathbb{R}^{W\times H\times C }$ over a frame $S_t$ from a video $V$ with width $W$, height $H$, number of channels $C$, and estimated optical flow field $\mathbf{o}_t$ between frames $ S_{t+1}\text{ and } S_{t}$. This short-term prior allows the network to explicitly assess the approximate appearance of the targets resulting in more precise segmentations.  

\vspace{-0.40cm}
\paragraph{Long-Term Attention Prior.}
We establish an LTA prior from the iterative estimation of an instance location. We calculate this prior by means of a bounding box tracker, which estimates an approximate location of an object at frame $t$ given its appearance and its immediate context at times $\{t-1,...,t-n\}$. This process is performed in a sequential manner, starting at frame 0, whose bounding box location information is known.  

Since tracking is, in essence, a one-shot learning problem \cite{bertinetto2016learning} and its core algorithm can be initialized and efficiently executed, even in large-scale datasets, it directly fits into the one-shot video instance segmentation framework. 

\vspace{-0.40cm}
\paragraph{Unit Optical Flow.}
Many recent methods for video object segmentation \cite{wang2017super,voigtlaender2017online,tokmakov2017learning,perazzi2017learning} focus on inferring pixel-wise mappings exclusively from spatial data. Optionally, these methods complement their predictions with temporal estimates over initial ground-truth labels. We depart from this mainstream approach and estimate explicit temporal cues from the video, which we fuse at frame-level with the spatial information.

We compute the optical flow from consecutive RGB frames using FlowNet2.0 \cite{dosovitskiy2015flownet} and map the estimated vector field $\mathbf{o}=(x,y)$ into its unitary direction field $\hat{\mathbf{o}}=\frac{o}{|o|}$. We concatenate these unitary vectors with the normalized magnitude of the flow field and name this 3-channel cue the \textit{Unit Optical Flow} (uOF). Compared to the raw vector field, the uOF is bounded between $[-1,1]$, thus it is more stable in the presence of large displacements and estimation errors. We use the uOF as a complementary source of information in MAIN, thereby achieving end-to-end training using both spatial and temporal cues. 

\vspace{-0.10cm}
\subsection{Multi-Scale Separable Decoder}
\label{subsec:MultiScaleEncoderDecoder}
\vspace{-0.10cm}
Since richer contextual information aids at better modeling local information and overall segmentation score, recent works \cite{du2015hierarchical, liu2015parsenet, zhao2017pyramid, chen2018deeplab, ronneberger2015u} establish connections between coarse and fine information to produce better segmentations. Our decoder uses multi-scale separable convolutional operators to augment its field-of-view. There are two elements at the core of this enhancement: dilated convolutions \cite{yu2015multi} that enables the exponential expansion of a neuron's receptive field without losing resolution \cite{yu2015multi, chen2018deeplab}; and separable convolutions that aim at factoring cross-channel and spatial correlations \cite{chollet2017xception}, reducing computation, increasing time efficiency and which we find to be beneficial for instance segmentation.

\vspace{-0.10cm}
\subsection{Implementation Details}
\vspace{-0.10cm}
We train MAIN using the Pytorch library \cite{paszke2017automatic}. We choose the ADAM optimizer \cite{kingma2014adam} and train until convergence with an initial learning rate of $1\times 10^{-4}$ and learning rate annealing $\gamma =0.1$ every 45000 iterations. The decoder layers are randomly initialized using the Xavier weight initialization \cite{glorot2010understanding}. To speedup training in the large-scale Youtube-VOS, which contains mostly frames of HD-quality at 1280$\times$768, we rescale the training data to 256$\times$416, leading to a batch-size of 23 in training using a Titan X GPU.


\vspace{-0.40cm}
\paragraph{Multi-Scale Separable Decoder.}
For our encoder backbone, we use Resnet50 pretrained on Image-Net \cite{deng2009imagenet}. We drop the final global pooling and fully connected layers and augment the backbone with side outputs \cite{hariharan2015hypercolumns, long2015fully} at the end of each block (the final layer before the pooling operator). We adopt the architectural pattern from Feature Pyramid Networks \cite{lin2017feature}, in which feature maps before every pooling operation are independently up-sampled by a factor of 2 and post-processed with a 4-layer feature pooling stack. The first layer of the stack is a standard 1x1 separable convolution and ReLU non-linearity \cite{krizhevsky2012imagenet}. The remaining 3 layers are 3$\times$3 separable convolutions  with dilation factors of 1, 2 and 3 \cite{yu2015multi}. 

 
The up-sample step is performed with a bilinear interpolation followed by a 3$\times$3 convolutional layer. The feature pooling and up-sampling process are performed independently over all the side-outputs of the encoder, thus, allowing incremental fusion (by an element-wise addition) of every up-sampled response map with the pooled feature map from the previous level.

%% file: sections/4Experiments.tex
\vspace{-0.10cm}
\section{Experimental Validation}
\vspace{-0.10cm}
In this section, we provide empirical evidence to support our approach. We proceed by validating each individual contribution and related design choice.

\vspace{-0.10cm}
\subsection{Datasets and Evaluation Metrics}
\vspace{-0.10cm}
\label{sec:datasets}
\paragraph{Youtube-VOS \cite{xu2018youtubeArxiv}.} It is the first large-scale video segmentation dataset consisting of 4453 high quality hand labeled videos, split across training (3471), validation (474) and test (508) sets. Following the one-shot formulation, this dataset provides the ground-truth segmentation for the first frame in any set. Xu. \textit{et al.} \cite{xu2018youtubeArxiv} reserve the test set for the annual challenge, and designate the validation set, whose ground truth labels are also withheld, for algorithm evaluation and ranking over a dedicated evaluation server.

Since annotations in the Youtube-VOS validation set are withheld \cite{xu2018youtubeArxiv}, we create a random split from the original training set. As a result, we obtain two subsets: \textit{Train66} (2314 videos) and \trainVal (1157 videos) with known labels and use these sets to empirically validate each of our contributions.


To assess the final performance of our MAIN method and to compare our results against the state-of-the-art, we follow the standard evaluation methodology of the Youtube-VOS benchmark introduced in \cite{xu2018youtubeArxiv}. The two main evaluation metrics are: region similarity or Jaccard index (\jaccard) and F-measure (\fmax) for contour accuracy and temporal stability. The Youtube-VOS validation set includes instances from categories not available in the training set, thus, creating two dataset-specific metrics for \textit{seen} and \textit{unseen} object categories. 


\vspace{-0.40cm}
\paragraph{DAVIS-17 \cite{pont20172017}.} We complement our empirical validation by assessing the effectiveness of MAIN on the DAVIS-17 dataset, which consists of 60 high quality videos for training and 30 for validation. This dataset is densely annotated but is two orders of magnitude smaller than Youtube-VOS. DAVIS-17 also evaluates \jaccard and \fmax but without the distinction between seen and unseen categories. 


\vspace{-0.10cm}
\subsection{Architectural Ablation Experiments}
\vspace{-0.10cm}
We now empirically assess our architectural design choices namely: dilated and separable convolutions in the decoder of our network. Table \ref{tab:architectureExp} summarizes the results. We train four versions of our network  with different combinations (absence/presence) of separable and dilated convolutions in \textit{Train66}. Evaluation in \trainVal shows that using separable and dilated convolutions is beneficial for our method.

\begin{table}[h!]
\small
\centering
\begin{tabular}{c|c|c}
\toprule
\textbf{Separable} & \textbf{Dilated} &\multirow{2}{*}{\textbf{\(\mathcal{J}\)} \textbf{Seen}}\\
\textbf{Convolutions} & \textbf{Convolutions} &\\
\midrule
\(\chi\)   &\(\chi\)   &	0.603 \\
\(\chi\)   &\checkmark &	0.606 \\
\checkmark &\(\chi\)   &	0.613 \\ 
\checkmark &\checkmark &	\textbf{0.645} \\\midrule
\end{tabular}
\caption{\textbf{Decoder architecture.} Comparison configurations of MAIN. We assess the performance of the different combinations of separable and dilated convolutions. Results presented on Youtube-VOS \trainVal set. Both Separable and Dilated Convolutions contribute to the performance of the method.
\vspace{-0.40cm}}
\label{tab:architectureExp}
\end{table}

\vspace{-0.40cm}
\paragraph{Loss Function:}
We evaluate the suitability of the proposed WID as loss function by comparing it against other loss functions. This comparative study is summarized in Table \ref{tab:losses}. We compare against Dice Coefficient and Binary Cross Entropy by training three versions of our network in \textit{Train66} with the different loss functions. WID clearly outperforms the other losses when evaluated in the \trainVal set.

\begin{table}[t]
\centering
\small
\begin{tabular}{l|c}
\toprule
\textbf{Loss Function} & \textbf{\(\mathcal{J}\)} \textbf{Seen}\\
\midrule
Binary Cross Entropy          & 0.538 \\ 
DICE                          & 0.611 \\ 
Weighted Instance DICE (WID)  & \textbf{0.645} \\ \midrule
\end{tabular}
\caption{\textbf{Loss functions.} Comparison of loss functions. We train MAIN with different loss functions and test them on Youtube-VOS \trainVal set. The proposed WID outperforms other loss function.
\vspace{-0.40cm}}
\label{tab:losses}
\end{table}

\vspace{-0.10cm}
\subsection{Attention Priors}
\vspace{-0.10cm}
To evaluate the effect of the different attention cues on MAIN, we ablate each attention cue to show their importance separately and jointly. Table \ref{tab:attentionExp} 
summarizes the ablation experiments.  We train eight different configurations of our network in \textit{Train66} by incrementally adding each of the different attention cues.  Each attention cue contributes to the performance of the network, with \LTA being especially useful at improving MAIN's performance, and \STA more suitable for refining results.

\begin{table}[b!]
\small
\centering
\begin{tabular}{l|l|c}
\toprule
\textbf{Attention} & \textbf{Input Data} & \textbf{\(\mathcal{J}\)} \textbf{Seen}\\
\midrule
None     & RGB     & 0.321 \\ 
None     & RGB+uOF & 0.326 \\ \midrule
STA      & RGB     & 0.436 \\ 
STA      & RGB+uOF & 0.500 \\ \midrule
LTA      & RGB     & 0.625 \\ 
LTA      & RGB+uOF & 0.628 \\ \midrule
LTA+STA  & RGB     & 0.632 \\ 
LTA+STA  & RGB+uOF & \textbf{0.645} \\ \midrule
\end{tabular}
\caption{\textbf{Attention cues.} Comparison of long term vs short term attention priors on Youtube-VOS \trainVal set. The proposed long term and short term cues enable the multi-instance segmentation task, and complement each other when used in conjunction.
\vspace{-0.40cm}}
\label{tab:attentionExp}
\end{table}
\vspace{-0.40cm}

\paragraph{Unit Optical Flow.}
We integrate the uOF by concatenating it with the RGB channels in the input. We initialize the first layer weights that correspond to the uOF inputs as the RGB average weights, thus adding $3$ channels to the input tensor. During training, we switch between the forward and backward estimates of the uOF. To verify the effectiveness of uOF compared to the raw optical flow, we train two versions of MAIN in \textit{Train66} switching this attention cue and evaluate them in \trainVal. Table \ref{tab:flows} shows that using uOF is more beneficial to MAIN than using raw optical flow. We also test the performance of the model when adding the uOF. Table \ref{tab:attentionExp} shows that for every configuration of attention cues the method benefits with the inclusion of uOF. 

\begin{table}[t]
\centering
\small
\begin{tabular}{l|c}
\toprule
\textbf{Optical Flow Configuration} & \textbf{\(\mathcal{J}\)} \textbf{Seen}\\
\midrule
Optical Flow        & 0.612 \\ 
Unit Optical Flow   & \textbf{0.645} \\ \midrule
\end{tabular}
\caption{\textbf{Robust optical flow.} Comparison of optical flow sources used as input for video segmentation. The Unit Optical Flow has a better performance compare to using the raw Optical Flow field.
\vspace{-0.40cm}}
\label{tab:flows}
\end{table}

\vspace{-0.40cm}
\paragraph{Multi-Instance Attention.}
\label{par:MultAttention}
We complement the input of the multi-scale encoder-decoder with a set of $2N$ attention maps ($N$ for LTA and $N$ for STA), in which $N$ is set to the maximum number of instances in the dataset (6 for Youtube-VOS and 10 for DAVIS-17). Each attention map encodes the estimated location of a single instance by means of a binary bounding box. For a video with $M$ instances ($M  <  N$), we set the remaining $N-M$ maps to $0$. We concatenate these additional $2N$ channels to the RGB and uOF input. Since this modification changes the input dimension, we initialize the first layer weights corresponding to the attention cues with the average of the original RGB weights, thus, avoiding the need to retrain the whole layer and favoring faster convergence. 

We evaluate the effectiveness of producing a multi-instance segmentation compared to estimating multiple single-instance predictions and then joining the results. Table \ref{tab:multiInstanceExp} shows that the multi-instance approach of MAIN significantly outperforms the single-instance scenario. 

\begin{table}[h!]
\centering
\small
\begin{tabular}{l|c}
\toprule
\textbf{Number of instances} & \textbf{\(\mathcal{J}\)} \textbf{Seen}\\
\midrule
Single-Instance     & 0.616 \\ 
Multi-Instance      & \textbf{0.645} \\ \midrule
\end{tabular}
\caption{\textbf{Multi-instance predicitions.} Comparison of multi-instance vs single-instance configuration on Youtube-VOS \trainVal set for MAIN. Multi-instance achieves a significantly better performance than single-instance.}
\vspace{-0.40cm}
\label{tab:multiInstanceExp}
\end{table}

\vspace{-0.40cm}
\paragraph{Long Term Attention Priors.}
We derive our \LTA from the tubes generated by the Distractor-Aware Siamese Tracker (DA-Siam-RPN) \cite{zhu2018distractor}. We keep the default anchor-ratios from the VOT-2018 dataset \cite{Kristan2018a} and set the number of candidate scales to 8, with a displacement penalty of 0.055 windowed by a cosine function, and template learning rate of 0.395. For the network weights, we use the 'AlexNet-Big' model. The tracker is initialized with a tight bounding box created over the instance annotation from the first frame and run over the full frame-set of Youtube-VOS. In the \textbf{supplementary material}, we show a comparison of the relative performance of the proposed attention methods according to the tracker's average overlap as the video progresses.

\LTA is one of the most important cues for MAIN. Table \ref{tab:attentionExp} shows the improvement of adding the LTA for each of MAIN configuration. We find it beneficial to initially train MAIN using the ground-truth segmentation to create perfect bounding boxes  and later, when the network converges, introduce the estimated LTA. This way, we allow the network to first learn to operate over attention cues and later to learn from error modes on the tracker. 

\vspace{-0.40cm}
\paragraph{Short Term Attention Priors.}
During the final training stage, we concatenate the STA attention to our input tensor. We extend the first layer weights, corresponding to the STA, by replicating the weights of the LTA input. Therefore, the final version of MAIN, has an input tensor of dimension $(6+2N) \times W \times H$ for the inputs: RGB, uOF, LTA, and STA. To train this last phase, we perform several data augmentation at different image scales and crops of size 256$\times$416. We replace the ideal LTA cue with the one estimated by the tracker, and for the STA at time $t$, we randomly choose the ground-truth segmentation from consecutive frames $[t-1,t+1]$. We randomly dilate and erode the selected annotation with squared kernels of sizes that vary between $6$ to $30$ pixels, perform affine transformations such as scale change between 0.8 and 1.2 of ratio, and shift between $0$ to $1\%$ of image size. All these transformations are used to approximate the possible errors that the previous segmentations might have in the validation set. In the forward phase, we simply take the previously segmented output and warp it to the current time position by using the backward estimated optical flow.


To adapt MAIN to the error pattern in the tracker, we use a Curriculum Learning  \cite{bengio2009curriculum} strategy. We steadily increase the error source (\textit{i.e}. tracker error) as the optimization process advances. We start by fine-tuning MAIN with LTA and STA data over the first 2, 4, 8 and 14 frames for three epochs each. Table \ref{tab:attentionExp} shows the significant improvement of adding the STA for each MAIN configuration. 

%% file: sections/5ComparisonSA.tex
\begin{table}[b]
\centering
\footnotesize
\begin{tabular}{ l | c | c | c | c | c }
\toprule
  \textbf{Method} & \textbf{OnT} & \textbf{\jaccard seen} & \textbf{\jaccard unseen} & \textbf{\fmax seen} &  \textbf{\fmax unseen} \\
\midrule
    S2S \cite{xu2018youtubeArxiv}  &\checkmark & \color{red}{0.710} & \color{red}{0.555} & \color{red}{0.700} & \color{blue}{0.612} \\
    \textbf{MAIN} & \(\chi\)& \color{blue}{\textbf{0.667}} &\color{blue}{\textbf{0.550}} & \color{blue}{\textbf{0.690}} & \color{red}{\textbf{0.630}}\\ 
    S2S \cite{xu2018youtubeArxiv} & \(\chi\)&  \color{blue}{0.667} & 0.482 & 0.655 & 0.503 \\
    OnAVOS \cite{voigtlaender2017online} &\checkmark & 0.601 & 0.466 & 0.627 & 0.514 \\
    OSMN \cite{yang2018efficient}        &\checkmark & 0.600 & 0.406 & 0.601 & 0.440 \\
    MaskTrack \cite{perazzi2017learning} &\checkmark & 0.599 & 0.450 & 0.595 & 0.479 \\
    OSVOS \cite{caelles2016one}          &\checkmark & 0.598 & 0.542 & 0.605 & 0.607 \\
\hline
\end{tabular}
\caption{ \textbf{Comparison of State-of-the-art methods} We evaluate MAIN on  Youtube-VOS validation set, scores are taken from \cite{xu2018youtubeArxiv}. 
For each metric we highlight in {\color{red}red} the best result and in {\color{blue}blue} the second best result. Results in \textbf{bold} correspond to our method. OnT is the abbreviation of Online Training.
\vspace{-0.10cm} 
}
\label{table:youtube}
\end{table}

\begin{table}[t]
\centering
\footnotesize
\begin{tabular}{ l | c | c | c }
\toprule
\textbf{Method} & \textbf{OnT} & \textbf{\jaccard} & \textbf{\fmax} \\
\midrule
    \textbf{MAIN}                  & \(\chi\) & \color{red}{\textbf{0.602}} & \color{red}{\textbf{0.657}}\\ 
    VideoMatch \cite{Hu_2018_ECCV} & \(\chi\) & \color{blue}{0.565} & - \\
    FAVOS \cite{cheng2018fast} & \(\chi\) & 0.546  & \color{blue}{0.618}\\
    OSMN  \cite{yang2018efficient} & \(\chi\) & 0.525  & 0.571\\
    SiamMask \cite{wang2018fast}   & \(\chi\) & 0.511  & 0.550\\
    MaskRNN \cite{hu2017maskrnn}   & \(\chi\) & 0.455  &  -   \\
    
    \midrule
    DyeNet \cite{Li_2018_ECCV} & \checkmark  & \color{red}{0.673} & \color{blue}{0.710}\\
    OnAVOS \cite{voigtlaender2017online} & \checkmark & \color{blue}{0.640} & \color{red}{0.712}\\
    VideoMatch \cite{Hu_2018_ECCV} & \checkmark & 0.614 & -\\
    OSMN \cite{yang2018efficient} & \checkmark & 0.608 & 0.571\\
    MaskRNN \cite{hu2017maskrnn} & \checkmark  & 0.605 &  -   \\
    OSVOS \cite{caelles2016one} & \checkmark & 0.566 & 0.639 \\
\hline
\end{tabular}
\caption{\textbf{DAVIS 17 benchmark.} Comparison of State-of-the-art methods on DAVIS Validation. Offline training methods are shown separated for a fair comparison. For each metric we highlight in {\color{red}red} the best result and in {\color{blue}blue} the second best result. Results in \textbf{bold} correspond to our method.
\vspace{-0.10cm}
}
\label{table:davis2017val}
\end{table}

\begin{figure}[b]
    \centering
    \includegraphics[trim={1.8cm 0.0cm 2.2cm 0.7cm}, width=0.73\linewidth]{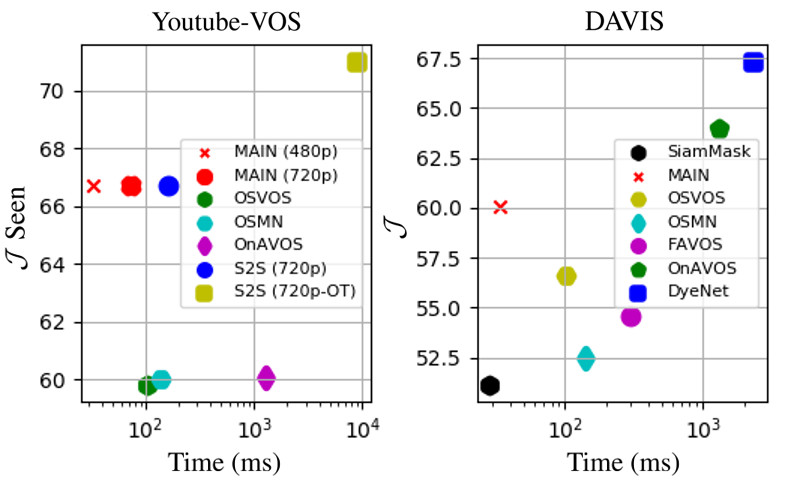}
    \caption{\textbf{Efficiency Benchmark.} Jaccard metric vs forward time (ms) on the Youtube-VOS and DAVIS 2017 datasets. Our method has a clear advantage on both benchmarks running on near real-time on DAVIS 2017.
    \vspace{-0.10cm}}
    \label{fig:speed}
\end{figure}

\vspace{-0.20cm}
\section{Comparison with the State-of-the-art}
\vspace{-0.10cm}

In this section, we compare our best MAIN network (RGB+uOF+LTA+STA) against state-of-the-art methods. Table \ref{table:youtube} summarizes this comparison for the task of instance video segmentation on the Youtube-VOS validation set. We follow the standard testing methodology proposed by \cite{xu2018youtubeArxiv}. We distinguish between offline and online methods for a fair comparison. As outlined in Section \ref{sec:datasets}, the Youtube-VOS dataset breaks up the evaluation metrics between seen and unseen categories. This distinction creates a large performance gap between both sets, the latter being far more difficult. 

\begin{figure*}[t!]
\centering
\includegraphics[width=0.95\textwidth]{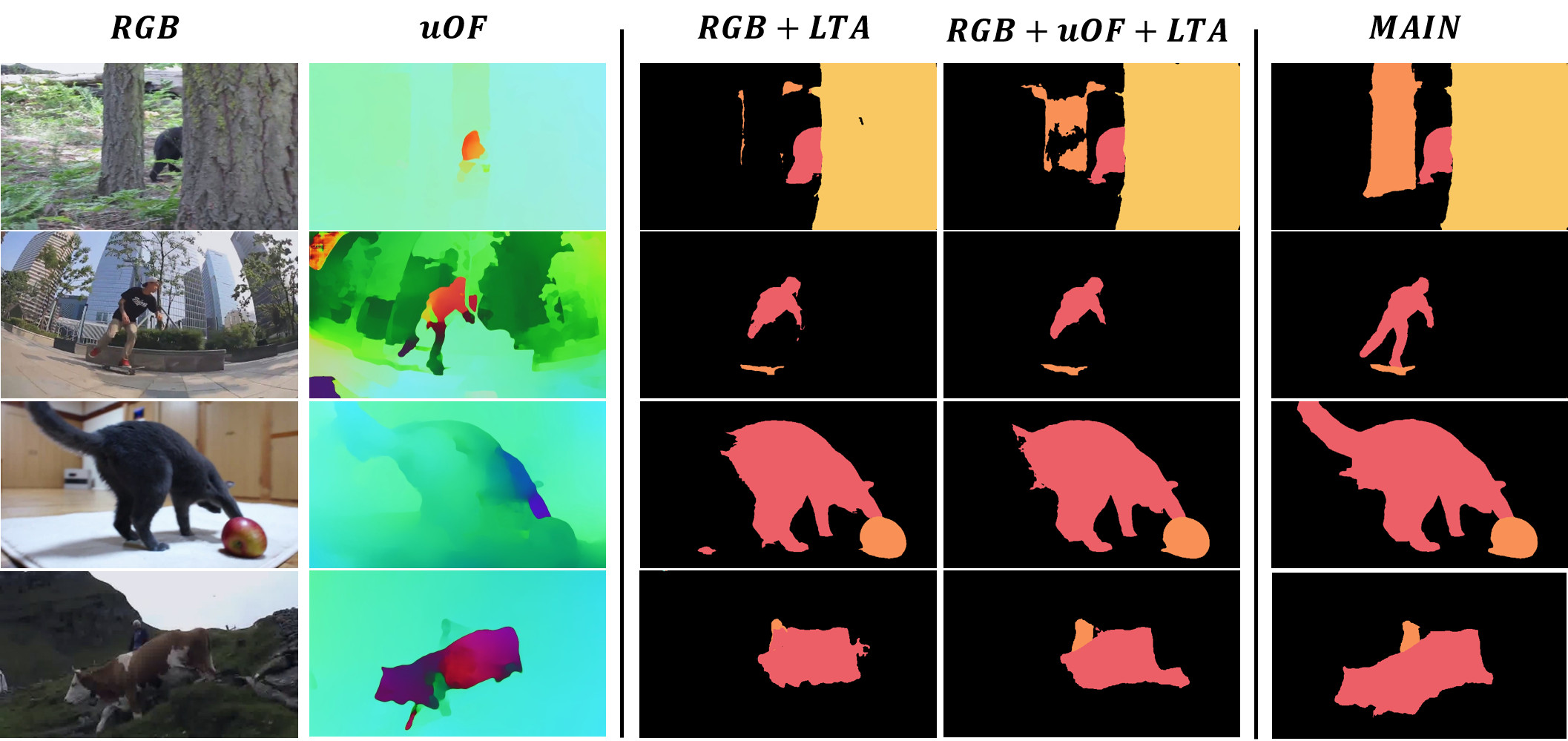}
\caption{\textbf{Qualitative results.} Each row represents a frame from a specific video. From left to right, the first column represents the original image. The second column is a representation of the Unitary Optical Flow. The third and fourth columns correspond to the results using only the configuration of $RGB+LTA$ and $RGB+uOF+LTA$ correspondingly. The last column corresponds to MAIN results.
\vspace{-0.40cm}}
\label{fig:vis}
\end{figure*}

MAIN achieves state-of-the-art scores for every metric in the offline configuration, achieving an accuracy that is competitive compared to online methods. Our method relies on generic attention cues and favors consistent temporal predictions along the video sequence, instead of enforcing strong semantics or individual appearance models. This leads to a significant performance improvement in the unseen scenario, in which MAIN surpasses the current offline state-of-the-art by $6.8\%$ and $12.7\%$ in \jaccard and \fmax metrics, respectively. Compared against the online methods, MAIN results are higher than almost all methods except for \cite{xu2018youtubeArxiv}, where MAIN is within  $0.5\%$ of the latter's performance in \jaccard unseen but outperforms it by $1.8\%$ in \fmax.

We also evaluate MAIN against the state-of-the-art on the DAVIS validation dataset, summarized in Table \ref{table:davis2017val}. We achieve state-of-the-art performance on the offline task, while remaining competitive against online methods. Furthermore, MAIN fares favorably against approaches that also rely on spatial attention like \cite{yang2018efficient}, and long-term recurrences like \cite{wang2018fast,hu2017maskrnn}.

\vspace{-0.10cm}
\subsection{Efficiency Analysis} 
\vspace{-0.10cm}
We conclude this section with an efficiency analysis. Figure \ref{fig:speed} shows the comparison between performance and inference time. We calculate the inference time for a single frame in MAIN after estimating the attention mechanisms. MAIN outperforms most state-of-the-art methods by an order of magnitude, where only SiamMask \cite{wang2018fast} performs faster in inference. It is important to emphasize that SiamMask was devised for fast inference and underperforms in \jaccard and \fmax when compared to MAIN. Additionally, SiamMask can only generate a single instance segmentation per forward pass. Hence, MAIN is faster than SiamMask in sequences with more than one instance.



%% file: sections/6QualitativeConclusions.tex
\vspace{-0.20cm}
\section{Qualitative Results}
\vspace{-0.10cm}
To complement our experimental validation, we present some qualitative results in Figure \ref{fig:vis} that show the visual differences of incrementally adding cues in training. On the first (top) row, MAIN overcomes a challenging scenario with occlusions, similar visual instances, and blur. The second row corresponds to a video with fast moving objects, different-sized objects, and complex motion patterns. In this case, motion and STA cues play a key role in the good performance because all instances tend to stay in the center, while the background is moving quickly. The third video shows a case with objects with distinct degrees of blur and with an incomplete label in the first frame (the tail of the cat is missing). Even though MAIN is not aware that cats have tails, it deduces it thanks to the attention mechanisms that ensure that the tail is part of the object. The final video (bottom) shows two segmentation targets one mostly occluded an the other completely visible and in motion. While the LTA and uOF approximate the location and number of instances, only the inclusion of STA approximates the targets shape.


Overall, there is an improvement when stacking different priors and sources of information. LTA mostly approximates the objects location but fails at providing a good recall, especially for finer details or highly textured objects. Using the uOF refines the prediction of the objects with somewhat homogeneous motion. Finally, incorporating STA  with LTA and uOF enhances the segmentation quality by reducing the false negatives and improving the boundary definition between adjacent masks. 
We include more examples in the \textbf{supplemental material} that show MAIN in diverse scenarios.

\vspace{-0.20cm}
\section{Conclusions}
\vspace{-0.10cm}
MAIN is an efficient approach devised for generic instance-level segmentation. It is trained end-to-end with a new WID loss suitable for class imbalance, generating multiple instance-level segmentations in a single forward pass. We validate MAIN in the first large-scale video segmentation benchmark, achieving state-of-the-art results while running at 30.3 FPS. The increments in the unseen metrics demonstrate MAIN's effectiveness at fusing generic grouping cues and producing temporally consistent segmentations without requiring domain-specific knowledge.


%% file: sections/7Acknowledgement.tex
\paragraph{Acknowledgement}
This work was partially supported by the King Abdullah University of Science and Technology (KAUST) Office of Sponsored Research, and by the German-Colombian Academic Cooperation between the German Research Foundation (DFG grant BR 3815/9-1) and Universidad de los Andes, Colombia.

%% file: sections/8Apendix.tex
\onecolumn
\section{Appendix}
\subsection{Temporal Consistency Analysis}
\begin{figure}[h]
    \small
    \centering
    \includegraphics[width=0.950\textwidth]{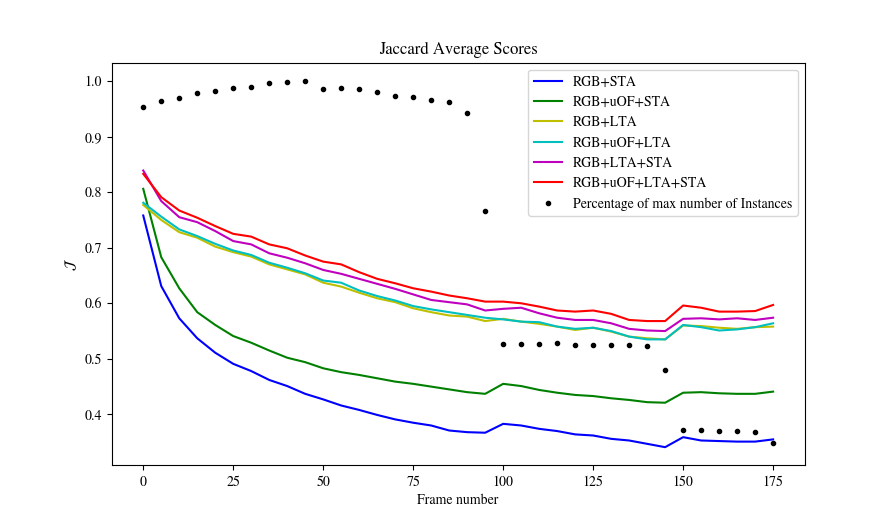}
    \caption{\textbf{Temporal Consistency.} Mean Jaccard scores along time according to the selected attention methods in the YoutubeVOS \trainVal set.}
    \label{fig:TemporalScores}
\end{figure}

To complement the experimental validation of section 4, we evaluate the temporal consistency of our method. Figure \ref{fig:TemporalScores} shows the performance of the different versions of our method (different choices for attention cues) as the video sequences progress. For this analysis we train on \textit{Train66} and validate over the \trainVal set.

Our method incrementally benefits with the addition of each attention prior. LTA is critical for MAIN's performance, if MAIN only considers STA the scores drop much faster during the first frames. We explain this behaviour as errors propagate faster if STA is the only source of attention. Overall, LTA improves the temporal consistency of predictions, enabling mask refinement by STA. Finally the combination of RGB, uOF and STA has a slower decrease than adding RGB+STA, which shows a complementary behaviour of the pixel-level and region-level short-term attention cues.

Like most video object segmentation methods, the performance of MAIN slowly decreases as the sequence progresses, it seems to stabilize around frame 100, above the 0.6 mean IoU. We will also note that the graph presents an unusual jump in performance around frames 100 and 150, specially for those configurations that only consider short-term priors. This particular behaviour is due to a bias in the videos length. In figure \ref{fig:TemporalScores} the black dots represents the number of instances still present at frame $n$ as fraction of the maximum number instances present at any time, the sharp declines in the number of instances correlate with the sudden jumps in the $\mathcal{J}$ of our methods, it seems that the remaining group of videos at frames 100 and 150 contain instances or scenarios that are easier to segment.

\subsection{Qualitative Results}
\begin{figure*}[h]
    \centering
    \includegraphics[width=1\textwidth]{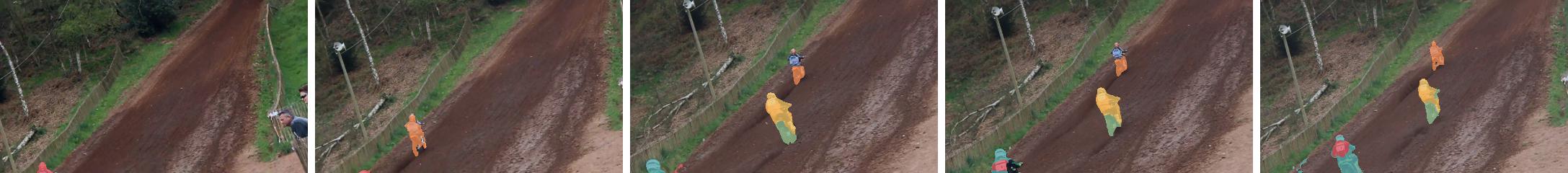}
    \caption{Shows a complex scenario where attention cues must be initialized from small, fast moving objects, namely three bikes and riders with very similar visual features and small size. While our algorithm manages to identify the objects of interest, it fails when the fast moving objects become smaller and generates overlapping predictions with false negatives. 
    }
    \label{fig:1bcd8a65de}
\end{figure*}

\begin{figure*}[h]
    \centering
    \includegraphics[width=1\textwidth]{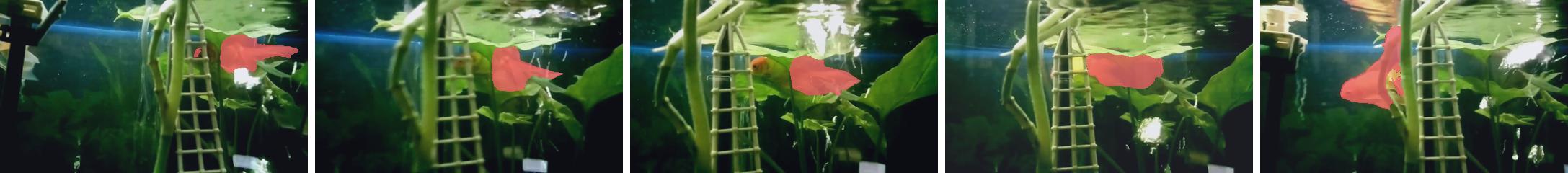}
    \caption{Illustrates an occlusion scenario that generates errors in the segmentation mostly due to the visual similarity of the segmentation target and its occluding background, along with the unusual lighting of this particular underwater scene. In this video MAIN propagates wrong estimations of the mask trough the sequence generating a fake segmentation mask around the occlusion boundary.}
    \label{fig:661f34feeb}
\end{figure*}

\begin{figure*}[h]
    \centering
    \includegraphics[width=1\textwidth]{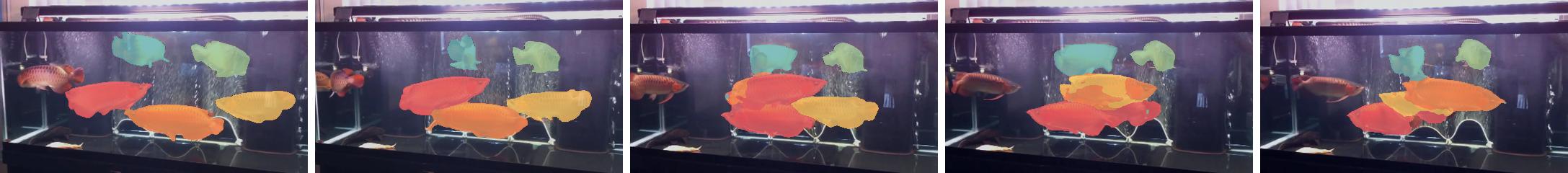}
    \caption{This is a hard scenario for MAIN, where visually similar instances overlap and occlude each other in complex patterns during a large period of time. While MAIN is still capable of identifying the target instances, it mixes their label information.}
    \label{fig:b1a8a404ad}
\end{figure*}

\begin{figure*}[h]
    \centering
    \includegraphics[width=1\textwidth]{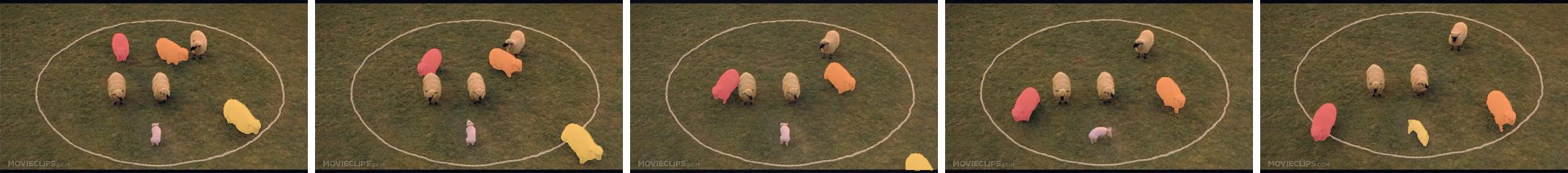}
    \caption{Presents an ideal segmentation scenario for MAIN, the instances appearance is almost unchanged trough the sequence, smooth motion patterns favor the propagation of attention cues and there is no overlap  between semantically similar objects in the scene. These conditions largely favor the accurate propagation of attention priors. This video is penalized mostly by errors at segmenting fine-grain details like the sheep's legs and face boundaries. }
    \label{fig:c74fc37224}
\end{figure*}

\begin{figure*}[h]
    \centering
    \includegraphics[width=1\textwidth]{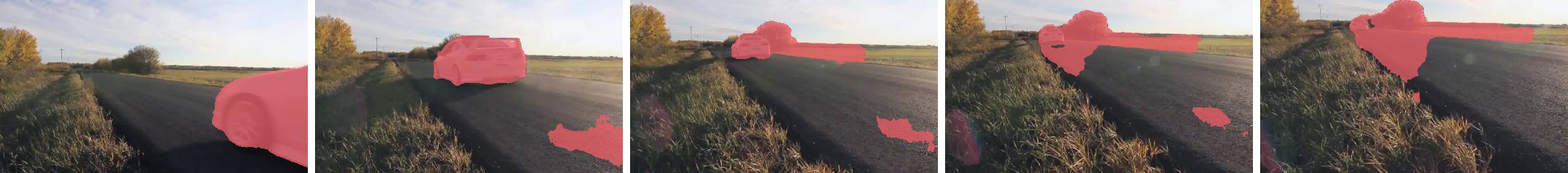}
    \caption{This video shows a relative simple scene with an single large object of interest. However, the size of the instance changes drastically along the video given its fast motion. This affects the estimation of all the priors leading to false positives. }
    \label{fig:f3678388a7}
\end{figure*}

\begin{figure*}[h]
    \centering
    \includegraphics[width=1\textwidth]{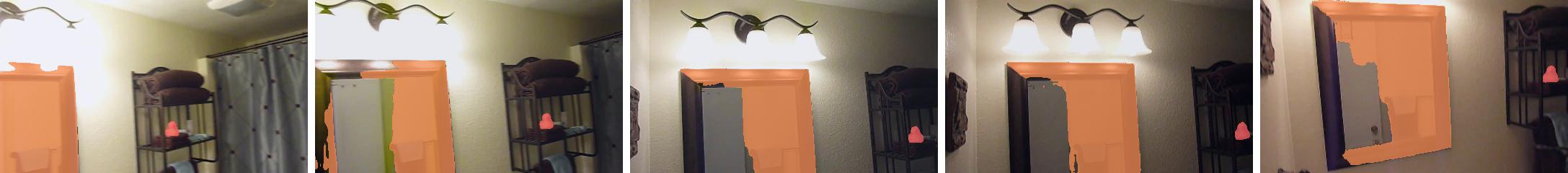}
    \caption{This video displays a scene with two objects of interest. The first one is a small object and the second one is a large mirror. Despite the small size of the first object, MAIN is capable to consistently segment it along the sequence. The mirror is a very hard segmentation target because of its constant change in 'appearance'. MAIN seems to attend to the reflected objects (a green door) and the mirror's frame.}
    \label{fig:4d6cd98941}
\end{figure*}

\begin{figure*}[h]
    \centering
    \includegraphics[width=1\textwidth]{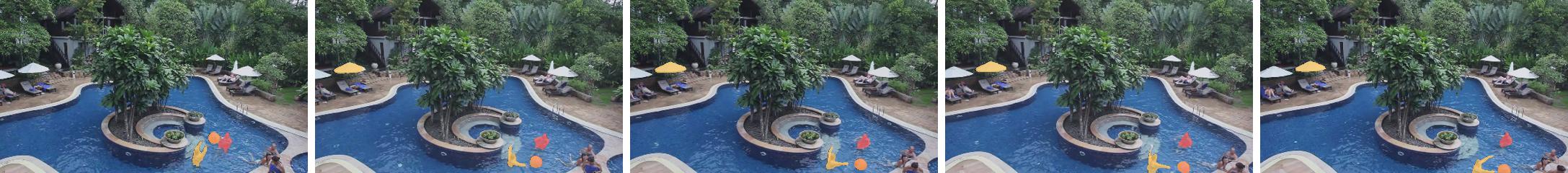}
    \caption{This figure visualizes three small objects that interact with each other. In this particular setting, the optical flow doesn't provide mayor clues, as the motion of the surrounding plants and the apparent motion of the water results in much larger local motion estimations than those of the segmentation targets. However, MAIN achieves a great segmentation over all three objects.}
    \label{fig:5f697fc636}
\end{figure*}

\begin{figure*}[h]
    \centering
    \includegraphics[width=1\textwidth]{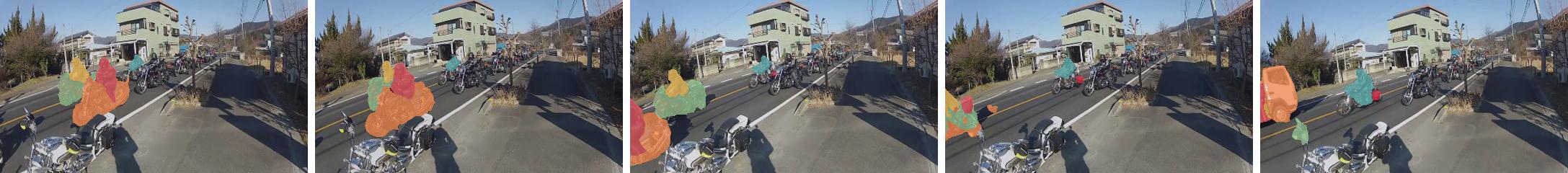}
    \caption{This scene shows five instances. Two of them are partially occluded. Nonetheless,  MAIN maintains a qualitatively acceptable segmentation until the last frames where it gets confused with the incoming truck.}
    \label{fig:5f8241a147}
\end{figure*}

\begin{figure*}[h]
    \centering
    \includegraphics[width=1\textwidth]{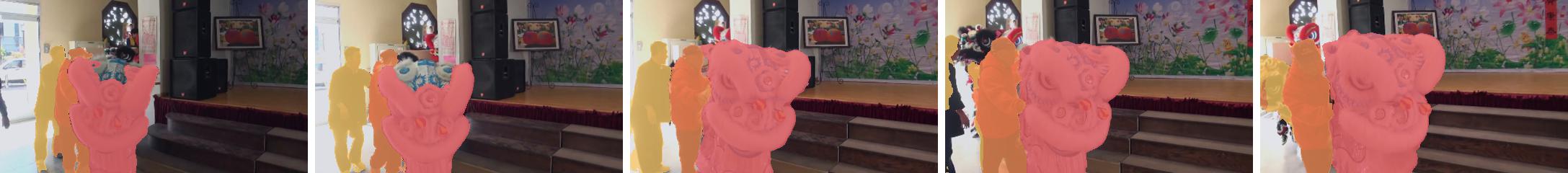}
    \caption{Evidences two very common instances  (humans), and an extremely unusual dragon-like figure. In this scenario, all three instances are segmented until the person in yellow disappears. It is remarkable that MAIN provides a mostly accurate segmentations of a complex instance whose appearance and scale change over time, correcting some initial false negatives}
    \label{fig:0788b4033d}
\end{figure*}

We provide more qualitative examples to better understand the capabilities and failure modes of MAIN. Figures \ref{fig:9c4419eb12} to \ref{fig:0788b4033d}  show challenging scenarios that include: occlusions (figures \ref{fig:9c4419eb12},  \ref{fig:13c3cea202}, \ref{fig:661f34feeb}, \ref{fig:b1a8a404ad}, \ref{fig:5f8241a147} and \ref{fig:0788b4033d}), scale changes (figures  \ref{fig:1bcd8a65de}, \ref{fig:d76ee8fa19},  \ref{fig:f3678388a7}), appearance changes (figures \ref{fig:13c3cea202}, \ref{fig:c74fc37224}, \ref{fig:d76ee8fa19} and \ref{fig:4d6cd98941}), multiple similar instances (figures \ref{fig:b1a8a404ad}, \ref{fig:c74fc37224} and \ref{fig:c74fc37224}), fast motion (figures  \ref{fig:1bcd8a65de}, \ref{fig:13c3cea202}, \ref{fig:d76ee8fa19} and \ref{fig:5f8241a147}) and very target small objects (\ref{fig:1bcd8a65de}, \ref{fig:4d6cd98941}, \ref{fig:5f697fc636}).

\begin{figure}[h]
    \centering
    \animategraphics[loop, autoplay,width=0.55\textwidth]{8}{suppl_animatedImages/9c4419eb12/image-}{0}{35}
    \caption{Exhibits three subjects and their Skiing equipment. This video provides a challenging scenario for MAIN due to the small size of the instances (see the  snowboard in yellow just below the person in orange). Additionally the person in green gets completely occluded during the sequence. MAIN is robust to false positive detections of the person in green, while still being accurate at segmenting the other small instances on the scene.  To visualize the animated figure use Adobe Acrobat Reader.}
    \label{fig:9c4419eb12}
\end{figure}

\begin{figure}[h]
    \centering
    \animategraphics[loop, autoplay,width=0.55\textwidth]{8}{suppl_animatedImages/13c3cea202/image-}{0}{19}
    \caption{Contains a difficult scenario for methods like MAIN that do not directly rely on semantics, here there is a large amount of overlap between the girl and the dog, moreover the quick motion patterns of the dog are a large source of  errors for all the attention cues. Nevertheless MAIN  manages to create an almost accurate boundary around the girl and has a high recall for the pixels in the dog instance. Its main source of error are wrong instance assignments between the girl and the dog, specially the girl's arm which has a similar color to the dog's fur.}
    \label{fig:13c3cea202}
\end{figure}

\begin{figure*}[h]
    \centering
    \animategraphics[loop, autoplay,width=0.55\textwidth]{8}{suppl_animatedImages/d76ee8fa19/image-}{0}{28}
    \caption{Shows an interesting scenario with a mirror that reflects the segmentation targets. There are minimal false positive detections located at the boundary between the small ape and its reflection. The main source of errors are erroneous instance label when the instances overlap. }
    \label{fig:d76ee8fa19}
\end{figure*}

